\documentclass[letterpaper, 10 pt, conference]{ieeeconf}

\IEEEoverridecommandlockouts
\usepackage{amsmath}
\usepackage{amssymb}
\usepackage{graphicx}
\usepackage{booktabs}
\usepackage[table]{xcolor}
\usepackage{url}
\usepackage{algorithm}
\usepackage{algpseudocode}
\usepackage{cite}

\graphicspath{{fig/}}
\usepackage[hidelinks]{hyperref}

\makeatletter
\let\DWG@citex\@citex
\def\DWG@firstcitekey#1,#2\@nil{#1}
\def\DWG@linkcitex[#1]#2{%
  \edef\DWG@ckey{\expandafter\DWG@firstcitekey\detokenize{#2},\@nil}%
  \hyperlink{cite.\DWG@ckey}{\kern-\fontdimen2\font\DWG@citex[#1]{#2}}%
}
\let\@citex\DWG@linkcitex
\let\DWG@bibitem\@bibitem
\def\@bibitem#1{\DWG@bibitem{#1}\Hy@raisedlink{\hypertarget{cite.#1}{}}}
\let\DWG@lbibitem\@lbibitem
\def\@lbibitem[#1]#2{\DWG@lbibitem[#1]{#2}\Hy@raisedlink{\hypertarget{cite.#2}{}}}
\makeatother

\newcommand{\runin}[1]{\par\noindent\textbf{#1.}\hspace{0.5em}\ignorespaces}

\title{\LARGE \bf
Drive2Gauss: FeedForward 4D Gaussian Reconstruction from Multimodal Generative Driving Videos
}

\author{Tong Zhao$^{1,2}$, Zhiyuan Han$^{2}$, Lin Chen$^{3}$, Bowen Xie$^{4}$\\
Yunxi Qiao$^{5}$, Bohan Li$^{6}$, Weiqing Xiao$^{7}$, Gen Li$^{8}$\\
Shu Han$^{9}$, Xuyang Dai$^{10}$, Cheng Bi$^{10}$, Hao Zhao$^{1}$,
Chaojian Li$^{11\dagger}$%
\thanks{$^{1}$Institute for AI Industry Research (AIR), Tsinghua University,
Beijing, China.}%
\thanks{$^{2}$Tongji University, Shanghai, China.}%
\thanks{$^{3}$Beijing Technology and Business University, Beijing, China.}%
\thanks{$^{4}$Fuzhou University, Fuzhou, China.}%
\thanks{$^{5}$Tsinghua University, Beijing, China.}%
\thanks{$^{6}$Shanghai Jiao Tong University, Shanghai, China.}%
\thanks{$^{7}$Nanjing University, Nanjing, China.}%
\thanks{$^{8}$Zhejiang University, Hangzhou, China.}%
\thanks{$^{9}$University of Wisconsin--Madison, Madison, WI, USA.}%
\thanks{$^{10}$Great Wall Motor, Baoding, China.}%
\thanks{$^{11}$The Hong Kong University of Science and Technology,
Hong Kong SAR, China.}%
\thanks{$^{\dagger}$Corresponding author.}%
}

\begin{document}

\maketitle
\thispagestyle{empty}
\pagestyle{empty}

\begin{abstract}
Generating reusable driving environments requires both controllable future observations and an explicit
representation that supports rendering from different viewpoints, yet existing methods still recover
explicit scenes from generated RGB videos through optimization for each sequence. To overcome these limitations, we present \textbf{Drive2Gauss}, a feedforward framework that converts jointly generated RGB, metric depth, and dynamic flow into dynamic 4D Gaussian scenes. A \textbf{Multimodal Video Generator} predicts the three
modalities from reference multiview images and future control signals. Furthermore, we introduce a \textbf{Gaussian Query Decoder}
that uses depth to initialize sparse 3D queries and flow to associate the queries across time, and
integrates appearance information through spatiotemporal feature aggregation and iterative query
refinement. The reconstruction stage operates in a single feedforward pass without optimization for each
scene. Extensive results on nuScenes demonstrate that our approach attains an FID of 6.38 and an FVD of 12.43 for future RGB generation, a PSNR of 29.91 dB when reconstructing its generated observations, and supports both direct rendering under lateral camera shifts and downstream video fixing. Code and more visualizations are available at \url{https://zttt04.github.io/Drive2Gauss_page/}.

\end{abstract}

\section{INTRODUCTION}

Controllable driving video generation produces future multiview observations conditioned on road layouts, object configurations, ego trajectories, and camera geometry~\cite{wen2023panaceapanoramiccontrollablevideo,gao2024magicdrivestreetviewgeneration,wang2024drivewm,wang2024drivedreamer,li2023drivingdiffusion,yang2024genad,gao2025magicdrivev2highresolutionlongvideo,hu2023gaia1generativeworldmodel,hu2024drivingworldconstructingworldmodel,chen2025unimlvgunifiedframeworkmultiview}. Such observations provide a source of synthetic driving data, complementary to the real-world logs in large-scale driving benchmarks~\cite{sun2020scalability,caesar2020nuscenesmultimodaldatasetautonomous,geiger2012kitti,alibeigi2023zod}, but a video alone does not define an explicit scene that can be rendered repeatedly from different viewpoints. Dynamic Gaussian representations offer this capability by encoding scene appearance, geometry, and motion effectively within a unified representation~\cite{kerbl20233dgaussiansplattingrealtime,wu2024fourdimensionalgaussian,yang2024deformablegaussians,yan2024streetgaussiansmodelingdynamic,zhou2024drivinggaussian,chen2025omnireomniurbanscene,hess2025splatad,zhou2024hugsholisticurban3d,peng2025desiregs4dstreetgaussians}.

\begin{figure}[t]
\centering
\includegraphics[width=\columnwidth]{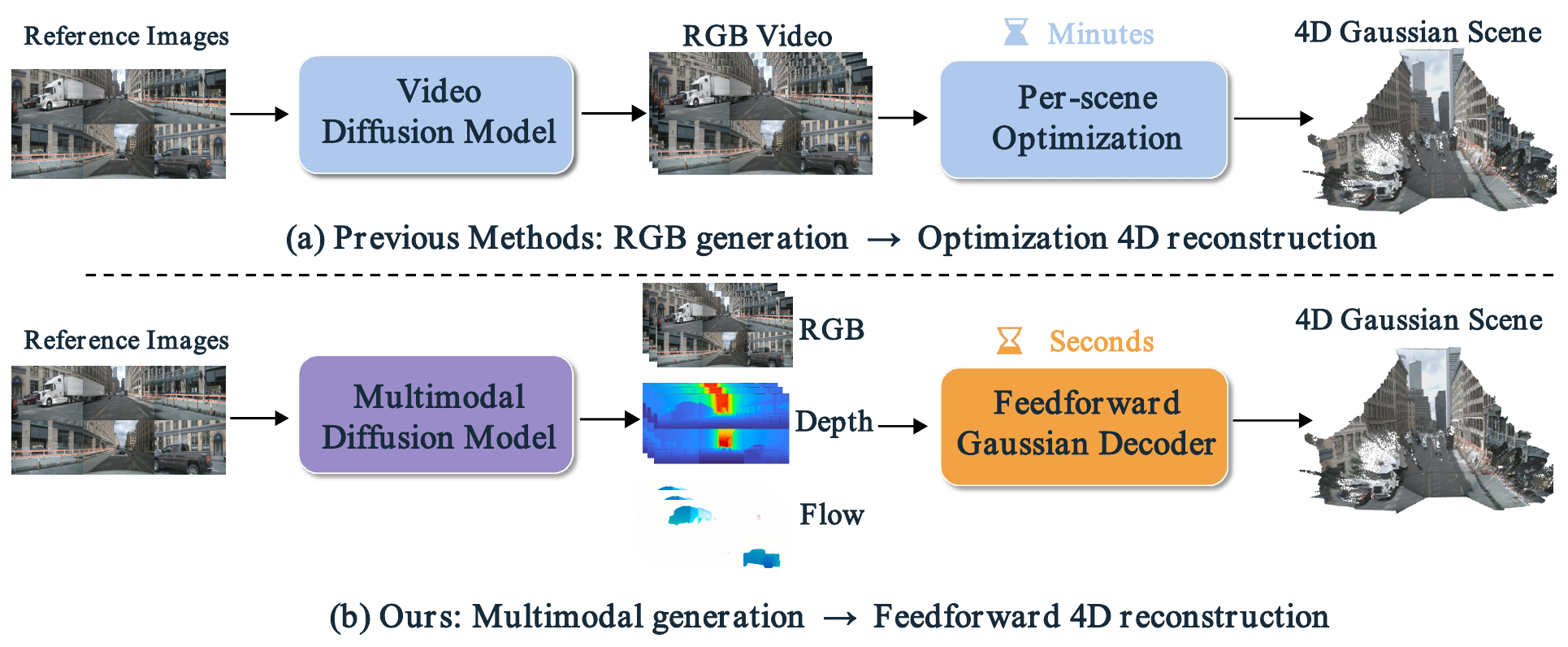}
\caption{Schematic comparison with RGB-first methods that use per-scene optimization. Drive2Gauss instead jointly generates RGB, metric depth, and dynamic flow, then constructs a 4D Gaussian scene with a feedforward decoder. The feedforward property refers to the reconstruction stage.}
\label{fig:method_comparison}
\end{figure}

Connecting generation with reconstruction requires geometric and temporal information that is consistent with the generated appearance. RGB-first pipelines, such as DriveDreamer4D, MagicDrive3D, DreamDrive, and ReconDreamer~\cite{zhao2025drivedreamer4d,gao2025magicdrive3dcontrollable3dgeneration,mao2025dreamdrivegenerative4dscene,ni2024recondreamercraftingworldmodels}, recover explicit scenes from generated observations through per-scene optimization. Multimodal generators provide an alternative source of structural information: DiST-4D generates RGB and metric depth, while other models incorporate occupancy and further scene modalities~\cite{guo2025dist4ddisentangledspatiotemporaldiffusion,li2025unisceneunifiedoccupancycentricdriving,omninwm2025,deng2026gaussiandwm,wang2024occsora4doccupancygeneration}. As illustrated in Fig.~\ref{fig:method_comparison}, existing generative reconstruction pipelines recover each sequence from generated RGB through per-scene optimization, whereas our approach directly reconstructs jointly generated RGB, depth, and flow with a single feedforward decoder.

Feedforward methods have also advanced explicit scene construction. Generalizable models infer geometry or Gaussian primitives from sparse views~\cite{Chen_2024,zhang2024gslrmlargereconstructionmodel,wang2025vggtvisualgeometrygrounded}, while DGGT and STORM recover dynamic scenes from driving images and PointForward aggregates multiview temporal features onto sparse 3D queries~\cite{dggt2025feedforward4dreconstruction,yang2024stormspatiotemporalreconstructionmodel,pointforward2026}. WorldSplat directly connects multimodal latent generation with a feedforward decoder that predicts pixel-aligned Gaussians~\cite{zhu2025worldsplatgaussiancentricfeedforward4d}. These developments motivate a more specific question: how can jointly generated depth and motion guide sparse query construction and temporal feature aggregation for temporally coherent explicit driving scenes?

We address this question with \textbf{Drive2Gauss}, a framework for feedforward generative driving scene reconstruction. Specifically, its \textbf{Multimodal Video Generator} jointly predicts RGB, metric depth, and dynamic flow. The \textbf{Gaussian Query Decoder} then assigns these modalities distinct roles: depth initializes 3D anchors, flow associates them across context frames, and RGB features supply appearance information. In this framework, cross-attention aggregates features from multiple views and context frames, and iterative query refinement updates the locations where features are sampled. The resulting queries are then decoded into dynamic Gaussian primitives without per-scene optimization. In contrast to WorldSplat~\cite{zhu2025worldsplatgaussiancentricfeedforward4d}, our decoder uses generated flow to establish
cross-frame correspondences. By drawing on jointly generated RGB,
depth, and flow, it also removes the need for the scene graph and
external depth estimator used by PointForward~\cite{pointforward2026}.

In summary, our main contributions are:
\begin{itemize}
\item We present Drive2Gauss, a framework that couples RGB, metric depth, and dynamic flow generation with feedforward 4D Gaussian reconstruction.
\item We develop a Gaussian Query Decoder that combines depth-based initialization, flow-guided temporal association, spatiotemporal feature aggregation, and iterative query refinement in a feedforward manner.
\item Extensive experiments on nuScenes~\cite{caesar2020nuscenesmultimodaldatasetautonomous} show that Drive2Gauss achieves state-of-the-art results in future video generation, feedforward 4D Gaussian reconstruction, and downstream view fixing, and the lowest FID under all evaluated lateral camera shifts.
\end{itemize}

\section{RELATED WORKS}

\subsection{Driving Video Generation}

Controllable driving generation has progressed from layout-conditioned multiview synthesis to predictive world models with longer horizons, broader scenario coverage, and richer control signals~\cite{li2023drivingdiffusion,wen2023panaceapanoramiccontrollablevideo,gao2024magicdrivestreetviewgeneration,wang2024drivedreamer,wang2024drivewm,yang2024genad,lu2024wovogen,gao2024vistageneralizabledrivingworld,gao2025magicdrivev2highresolutionlongvideo,zhao2025drivedreamer2,hu2023gaia1generativeworldmodel,hu2024drivingworldconstructingworldmodel,guo2024infinitydrivebreakingtimelimits,chen2025unimlvgunifiedframeworkmultiview,wang2024stag1realistic4ddriving,huang2024subjectdrivescalinggenerativedata,nvidia2025cosmostransfer1conditionalworldgeneration,ren2025gen3c3dinformedworldconsistentvideo,lu2025infinicubeunboundedcontrollabledynamic,jiang2024diveditbasedvideogeneration,ji2025cogen3dconsistentvideo}. These methods generate future observations or intermediate features for simulation and planning, but their outputs remain view-dependent rather than a persistent scene representation.

Beyond RGB, several driving generators add complementary scene signals: DiST-4D predicts metric depth together with RGB~\cite{guo2025dist4ddisentangledspatiotemporaldiffusion}, occupancy-centric generators synthesize 3D scene structure directly~\cite{li2025unisceneunifiedoccupancycentricdriving,wang2024occsora4doccupancygeneration,li2026scalingoccupancycentricdrivingscene,yan2025drivingspherebuilding}, and occupancy can also be estimated from images with Gaussian splatting~\cite{gan2025gaussianocc}. Others instead reason jointly over multimodal 4D observations~\cite{liu2025omnisceneattentionaugmentedmultimodal4d}, synthesize views along flexible camera poses~\cite{wang2024freevsgenerativeviewsynthesis}, or unify scene understanding and multimodal generation~\cite{omninwm2025,deng2026gaussiandwm}; comparable modeling appears outside driving as well, in generic video generation~\cite{geovideo2025} and embodied manipulation~\cite{rynnworld2026}. A further line forecasts future 3D occupancy, either toward planning~\cite{zheng2024occworldlearning3doccupancy,yang2025drivingoccupancyworld} or as a pretext task for 4D scene understanding~\cite{min2024driveworld4dpretrained}. Drive2Gauss uses jointly generated RGB, depth, and flow as complementary inputs for explicit 4D reconstruction instead of producing only view-dependent observations.

\subsection{Driving Scene Reconstruction}

Neural radiance fields and Gaussian splatting provide continuous or explicit representations for novel-view rendering~\cite{mildenhall2020nerfrepresentingscenesneural,barron2021mipnerfmultiscalerepresentationantialiasing,kerbl20233dgaussiansplattingrealtime,liu2024citygaussianrealtime}. Their dynamic extensions model temporal deformation, while driving-specific methods decompose static backgrounds, moving objects, semantics, and motion fields in complex urban scenes~\cite{wu2024fourdimensionalgaussian,yang2024deformablegaussians,zhou2024drivinggaussian,yan2024streetgaussiansmodelingdynamic,chen2025omnireomniurbanscene,zhou2024hugsholisticurban3d,peng2025desiregs4dstreetgaussians,chen2025periodicvibrationgaussiandynamic,hess2025splatad,turki2023suds,xia2025d2gs,kung2025lihigs,wang2025appearance}; several systems further extend such reconstructions to simulation, extrapolation, and editable assets~\cite{chen2021geosim,wu2024mars,hwang2024vegs,du2025rgegs}. These approaches typically optimize a separate representation for each observed sequence. Generative reconstruction methods similarly convert synthesized observations into explicit scenes through per-scene optimization~\cite{zhao2025drivedreamer4d,gao2025magicdrive3dcontrollable3dgeneration,mao2025dreamdrivegenerative4dscene,ni2024recondreamercraftingworldmodels}.

Feedforward reconstruction methods infer geometry or Gaussian representations directly from input observations~\cite{charatan2024pixelsplat3dgaussiansplats,Chen_2024,zhang2024gslrmlargereconstructionmodel,wang2025vggtvisualgeometrygrounded,wang2025continuous3dperceptionmodel,tian2024drivingforwardfeedforward3dgaussian,yang2024stormspatiotemporalreconstructionmodel,dggt2025feedforward4dreconstruction,recondrive,jiang2025anysplatfeedforward3dgaussian,ren2026tokengsdecoupling3dgaussian,lis2026tokensplat,zhang2026d4rt}. Generative approaches instead synthesize structured 3D or dynamic scene representations~\cite{yang2025prometheus3dawarelatentdiffusion,pan2026diff4splatcontrollable4dscene,xiang2024trellis}, and diffusion-based refinement further improves the rendered observations~\cite{wu2025difix3dimproving3dreconstructions}. Temporal correspondence models further provide persistent point tracks or geometry across video frames~\cite{xiao2024spatialtracker,zhang2025tapip3d}. Among the closest methods, WorldSplat~\cite{zhu2025worldsplatgaussiancentricfeedforward4d} decodes multimodal latents into pixel-aligned Gaussians, while PointForward~\cite{pointforward2026} initializes sparse 3D queries from external depth and associates dynamic queries through scene-graph instance transformations. Drive2Gauss instead obtains both cues from its Multimodal Video Generator: generated depth initializes 3D queries, and generated flow associates corresponding observations across context frames for feature aggregation.

\section{Method}

\begin{figure*}[t]
\centering
\includegraphics[width=0.9\textwidth]{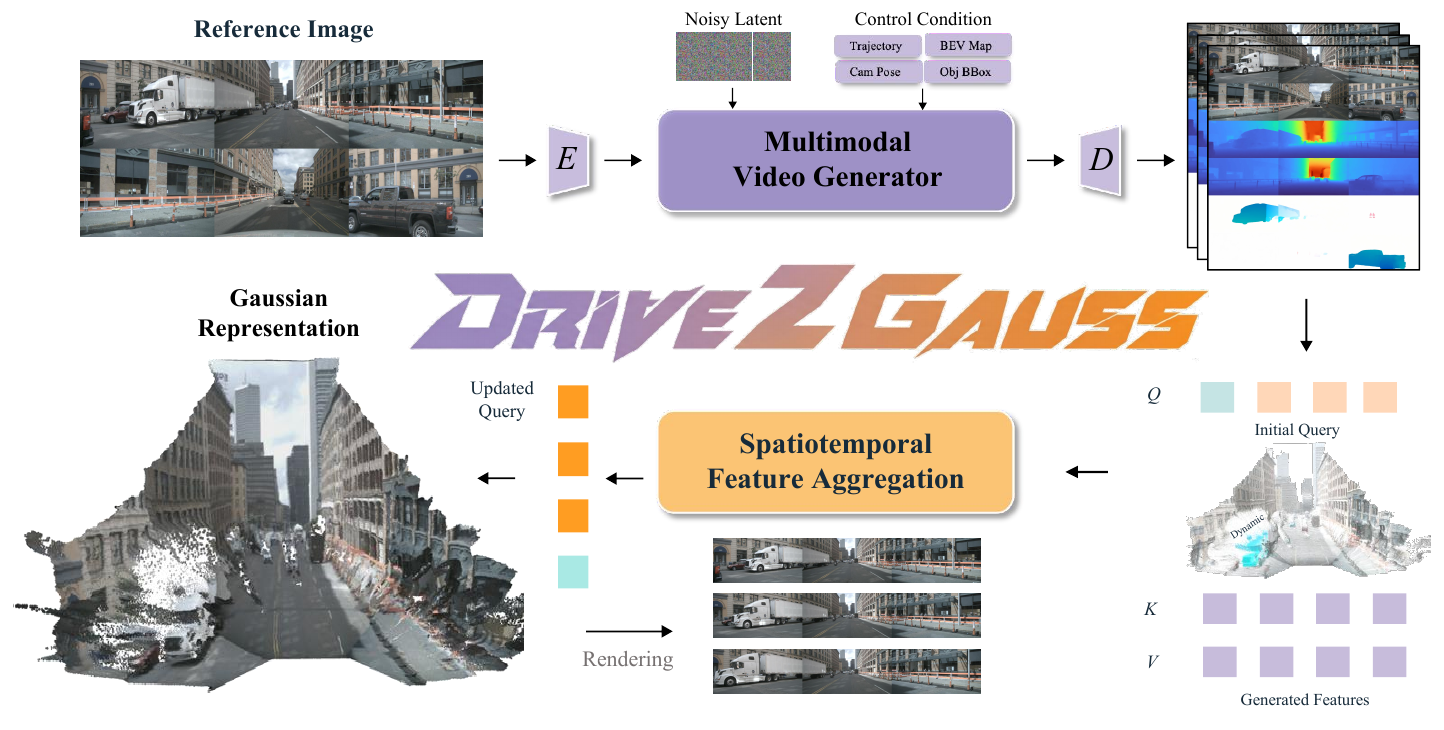}
\caption{Overview of Drive2Gauss. Reference multiview images and control
conditions are processed by the Multimodal Video Generator to predict
RGB, metric depth, and dynamic flow. Depth and flow initialize 3D queries,
while RGB features from multiple views and context frames are aggregated by the
Spatiotemporal Feature Aggregation module. The updated queries are decoded
into dynamic Gaussian primitives for rendering.}
\label{fig:method_main}
\end{figure*}

Our goal is to construct a future dynamic 4D Gaussian scene from a given reference frame of multi-camera observations. To this end, Drive2Gauss consists of two key components: a Multimodal Video Generator and a Gaussian Query Decoder. Given the reference observations and control signals, the generator jointly predicts future RGB, depth, and dynamic flow, providing complementary appearance, geometric, and motion cues for reconstruction. The decoder then uses these multimodal observations to initialize queries, aggregate spatiotemporal features, and decode them into dynamic Gaussian primitives in a single feedforward pass. An overview of our framework is illustrated in Fig.~\ref{fig:method_main}.

\subsection{Multimodal Future Video Generation}
\label{sec:multimodal_generation}

\runin{Joint Multimodal Generation}
Given multi-camera reference observations and the corresponding control signals, the Multimodal Video Generator jointly predicts future RGB, metric depth, and dynamic flow for all camera views. Let
\begin{equation}
\mathbf{I}_{\mathrm{ref}}
=
\left\{
\mathbf{I}_{\mathrm{ref},c}
\right\}_{c=1}^{C},
\end{equation}
where $C$ is the number of camera views and $\mathbf{I}_{\mathrm{ref},c}$ is the reference frame of view $c$.
For each of the $T$ future timestamps $t$, we define the control signal as:
\begin{equation}
\mathbf{S}_t
=
\left(
\mathbf{M}_t,
\mathbf{B}_t,
\mathbf{A}_t,
\mathbf{P}_t
\right),
\end{equation}
where $\mathbf{M}_t$ denotes the road map representation, $\mathbf{B}_t$ denotes the 3D bounding boxes of traffic participants, $\mathbf{A}_t$ describes the ego trajectory, and $\mathbf{P}_t$ denotes the camera geometry, including the intrinsic matrices and extrinsic poses for each view. Given the reference observations and the sequence of future control signals $\{\mathbf{S}_t\}_{t=1}^{T}$, the generator predicts
\begin{equation}
\left\{
\left(
\mathbf{Z}^{\mathrm{I}}_t,
\mathbf{Z}^{\mathrm{D}}_t,
\mathbf{Z}^{\mathrm{F}}_t
\right)
\right\}_{t=1}^{T}
=
G_{\theta}
\left(
\mathbf{I}_{\mathrm{ref}},
\left\{
\mathbf{S}_t
\right\}_{t=1}^{T}
\right),
\end{equation}
where $G_\theta$ denotes the video generator model with parameters $\theta$, and $\mathbf{Z}^{\mathrm{I}}_t$, $\mathbf{Z}^{\mathrm{D}}_t$, and
$\mathbf{Z}^{\mathrm{F}}_t$ denote the latent representations of RGB, metric depth, and dynamic flow at the future timestamp $t$, respectively. Decoding these latent representations yields the corresponding future RGB, depth, and dynamic flow observations for each individual camera view.

The generator operates in a compressed spatiotemporal latent space to support joint multimodal generation across multiple camera views~\cite{blattmann2023stablevideodiffusionscaling}. A pretrained 3D VAE~\cite{yang2025cogvideoxtexttovideodiffusionmodels} separately encodes RGB, metric depth, and dynamic flow into 16-channel latent features for each individual modality. The resulting features are concatenated along the channel dimension to form a unified 48-channel representation, which is processed by an STDiT denoising backbone with a ControlNet-style control pathway~\cite{zheng2024opensorademocratizingefficientvideo,zhang2023addingconditionalcontroltexttoimage}. Reference observations and map features are injected into the control pathway, while 3D bounding boxes, ego trajectories, and camera poses serve as conditioning inputs for both the main and control branches, following the same conditioning design of~\cite{gao2025magicdrivev2highresolutionlongvideo}. 

\runin{Geometric and Motion Supervision} For each training clip, we construct metric depth and dynamic flow as
complementary geometric and motion cues for 4D Gaussian reconstruction.
The constructed signals serve as supervision targets for a multimodal
video generation model, which learns to predict the corresponding modalities
for future video frames.

\emph{Metric depth.}
We follow the standard metric depth curation pipeline used in~\cite{guo2025dist4ddisentangledspatiotemporaldiffusion}, which combines
stereo depth from multiple views~\cite{ding2022transmvsnet} and temporally aggregated LiDAR
data, with dynamic objects and sky regions removed beforehand using
semantic segmentation~\cite{jain2023oneformer}. We then apply generative depth completion to
obtain dense metric depth maps for all training frames~\cite{liu2024depthlab}.

\emph{Dynamic flow.}
We first estimate dense optical flow between adjacent RGB frames using
SEA-RAFT~\cite{wang2024sea}. Since the raw optical flow also contains motion
caused by the ego vehicle and camera movement, we construct a dynamic mask
from the annotated 3D bounding boxes. Objects whose annotated velocity
exceeds a predefined threshold are treated as dynamic. Their 3D boxes are
projected onto the image plane and spatially dilated, and SAM-based instance
masks~\cite{kirillov2023segment} are used to refine the object boundaries. The dynamic flow is then
obtained by masking the optical flow:
\begin{equation}
\mathbf{F}^{\mathrm{dyn}}_t
=
\mathbf{M}^{\mathrm{dyn}}_t
\odot
\mathbf{F}^{\mathrm{opt}}_t,
\end{equation}
where $\mathbf{F}^{\mathrm{opt}}_t$ denotes the SEA-RAFT optical flow and
$\mathbf{M}^{\mathrm{dyn}}_t$ denotes the refined binary dynamic mask. Flow values
outside the dynamic regions are assigned a neutral value and marked as invalid.
This masked optical flow retains image correspondences within dynamic regions and effectively suppresses background flow. The target-frame depth and camera geometry are subsequently used to recover the associated positions in a common 3D reference coordinate system.

\subsection{Gaussian Query Decoder}
\label{sec:query_decoder}
\runin{Query Initialization} Following the point-aligned, query-based design of PointForward~\cite{pointforward2026}, we initialize 3D queries from the generated metric depth, and associate them across frames with the generated dynamic flow instead of an external depth estimator and scene-graph instance transformations. We sample a fixed number of queries in each camera view and context frame (the
four consecutive frames of a reconstruction window), keeping only those pixels with a valid depth value. Query $i$ is initialized from the pixel $(u_i,v_i)$ of view $c_i$ at timestamp $t_i$, with $d_i$ its depth along the corresponding optical axis. We unproject it into 3D and transform it into the reference frame:
\begin{equation}
\mathbf{x}^{0}_i
=
\mathbf{T}_{t_i\rightarrow r}
\mathbf{T}_{c_i\rightarrow l}
\left[
d_i\mathbf{K}_{c_i}^{-1}
\begin{pmatrix}
u_i\\v_i\\1
\end{pmatrix}
\right],
\end{equation}
where $\mathbf{K}_{c_i}$ denotes the intrinsic matrix of view $c_i$, $l$ is the ego frame at timestamp $t_i$, $r$ is the common reference frame, and $\mathbf{T}_{c_i\rightarrow l}$ and $\mathbf{T}_{t_i\rightarrow r}$ are the corresponding camera and temporal transformations. We also compute the ray direction $\mathbf{r}_i$ and Pl\"ucker moment $\mathbf{m}_i$ in the same reference frame, and sample the corresponding RGB value $\mathbf{I}_i$ at the same pixel.

To capture temporal motion, we use the generated dynamic flow to associate each query
across the surrounding context frames. For a given query from frame $s$, the flow displacement $\Delta\mathbf{p}_i^{f}$ at
its source pixel $(u_i^s,v_i^s)=(u_i,v_i)$ gives the corresponding tracked location in another
context frame $f$:
\begin{equation}
(u_i^f,v_i^f)
=
(u_i^s,v_i^s)
+
\Delta \mathbf{p}_i^{f} .
\end{equation}
The tracked location is unprojected using the corresponding target frame depth and transformed into the same reference coordinate system as the source point. Denoting the source and target positions by $\mathbf{x}_i^{s}$ and $\mathbf{x}_i^{f}$, respectively, the temporal anchor of query $i$ in the target frame $f$ is obtained as:
\begin{equation}
\bar{\mathbf{x}}^{f}_i
=
\mathbf{x}_i^{s}
+
p_i^{\mathrm{flow}}
\operatorname{clip}_{d_{\max}|f-s|}
\left(
\mathbf{x}_i^{f}-\mathbf{x}_i^{s}
\right),
\end{equation}
where $p_i^{\mathrm{flow}}$ is the dynamic probability from the flow, $d_{\max}$
is the maximum displacement per frame, and
$\operatorname{clip}_{r}(\mathbf{v})=\mathbf{v}\min(1,\,r/\|\mathbf{v}\|)$
limits a displacement to length $r$. Thus a query with zero probability keeps
its source position.

In summary, the query state is finally formed by combining the normalized 3D anchor, the source appearance, the camera ray information, and the corresponding motion prior:
\begin{equation}
\mathbf{q}_i
=
\left[
\operatorname{Norm}(\mathbf{x}^{0}_i),
\mathbf{I}_i,
\mathbf{r}_i,
\mathbf{m}_i,
p_i^{\mathrm{flow}}
\right].
\end{equation}
where $\operatorname{Norm}$ standardizes the anchor coordinates, and
$\mathbf{m}_i=\mathbf{o}_{c_i}\times\mathbf{r}_i$ is the Pl\"ucker moment of the
given unit ray $\mathbf{r}_i$, with $\mathbf{o}_{c_i}$ the camera center position in the reference frame.

\begin{figure}[t]
\centering
\includegraphics[width=0.72\columnwidth]{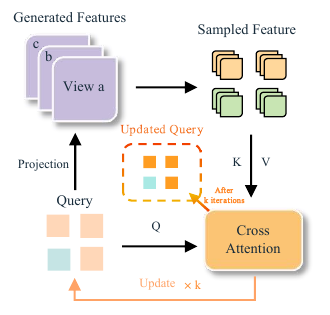}
\caption{Details of the spatiotemporal feature aggregation. Each query is
projected into valid camera views and context frames to sample features from
the generated observations, along with the corresponding key and value
features. Cross-attention (Q, K, V) fuses the sampled features to update the current
query location, and the update is repeated $k$ times, after which the refined
queries are passed to the Gaussian decoding stage for rendering.}
\label{fig:aggregation_detail}
\end{figure}

\runin{Spatiotemporal Feature Aggregation}Given the initialized and temporally associated queries, we aggregate visual evidence from multiple camera views and context frames to refine both query features and spatial locations (Fig.~\ref{fig:aggregation_detail}).
For each query, its temporal anchor is projected into every valid context observation, and the corresponding feature is bilinearly sampled from the Turbo-VAED decoder feature map~\cite{zou2025turbovaedfaststabletransfer}.

The sampled features are then aggregated through cross attention~\cite{vaswani2017attention}, where the query representation serves as the Query and the sampled multiview temporal features serve as the corresponding Keys and Values. Projections outside the image boundaries or associated with invalid tracks are excluded using a validity mask. At aggregation iteration $j$, a position head predicts a spatial offset $\Delta\mathbf{x}^{j}_i$ from the aggregated query representation and updates the current position of query $i$ iteratively, starting from the initial position $\mathbf{x}^{0}_i$:
\begin{equation}
\mathbf{x}^{j+1}_i
=
\mathbf{x}^{j}_i
+
\Delta \mathbf{x}^{j}_i.
\end{equation}
The updated query positions are then reprojected to resample visual features, allowing feature aggregation and query refinement to proceed iteratively. This procedure is repeated $k$ times, after which the refined query representations and 3D positions are passed to the Gaussian decoding stage.

\runin{Gaussian Decoding and Rendering}
After spatiotemporal feature aggregation, each refined query contains an updated feature representation and a corresponding 3D position. We decode each query into a dynamic Gaussian primitive by predicting its spatial offset $\Delta\boldsymbol{\mu}_i$, scale $\mathbf{s}_i$, rotation $\boldsymbol{\rho}_i$, opacity $o_i$, appearance feature $\mathbf{a}_i$, and temporal lifespan $\ell_i$. The Gaussian center at timestamp $t$ is defined as:
\begin{equation}
\boldsymbol{\mu}_i(t)
=
\mathbf{x}_i(t)
+
\Delta\boldsymbol{\mu}_i,
\end{equation}
where $\mathbf{x}_i(t)$ is the temporal anchor of query $i$ at timestamp $t$,
obtained by interpolating the tracked anchors of nearby context frames, and
$\Delta\boldsymbol{\mu}_i$ is the decoded spatial correction. The lifespan parameter $\ell_i$ controls the temporal visibility of
each Gaussian, allowing static queries to remain stable while dynamic queries
follow their estimated motion.

For rendering at timestamp $t'$ and camera view $c$, the Gaussian primitives are projected using the corresponding camera geometry and rasterized with a differentiable Gaussian renderer~\cite{kerbl20233dgaussiansplattingrealtime}. Following SplatAD and UniSim~\cite{hess2025splatad,yang2023unisim}, each Gaussian stores a high-dimensional appearance feature representation rather than directly predicting the RGB color. The rendered feature map is finally decoded by a lightweight appearance UNet to produce the final RGB image outputs.

\runin{Training Loss}
The Gaussian Query Decoder is supervised directly in image space: the predicted Gaussians are rasterized, the appearance UNet converts the rendered features to RGB, and these images are compared with the target RGB observations. The decoder objective is defined as follows:
\begin{equation}
\mathcal{L}
=
\mathcal{L}_{\mathrm{rgb}}
+
\lambda_{p}\mathcal{L}_{\mathrm{LPIPS}}
+
\lambda_{o}\mathcal{L}_{\mathrm{opacity}}
+
\lambda_{s}\mathcal{L}_{\mathrm{scale}},
\end{equation}
where $\mathcal{L}_{\mathrm{rgb}}$ is the Charbonnier loss for pixel reconstruction and $\mathcal{L}_{\mathrm{LPIPS}}$ measures perceptual similarity~\cite{zhang2018deepfeatures}. The terms $\mathcal{L}_{\mathrm{opacity}}$ and $\mathcal{L}_{\mathrm{scale}}$
regularize Gaussian opacity and scale, with weights $\lambda_o$ and $\lambda_s$;
$\lambda_p$ weights the perceptual term.

\section{EXPERIMENTS AND RESULTS}
\definecolor{bestcell}{RGB}{232,255,194}
\definecolor{secondcell}{RGB}{255,221,190}
\newcommand{\bestcell}[1]{\cellcolor{bestcell}\textbf{#1}}
\newcommand{\secondcell}[1]{\cellcolor{secondcell}{#1}}
\newcommand{\bestlabel}{\colorbox{bestcell}{\textbf{best}}}
\newcommand{\secondlabel}{\colorbox{secondcell}{{second}}}
\newcommand{\cmark}{\textcolor{green!60!black}{$\checkmark$}}
\newcommand{\xmark}{\textcolor{red!80!black}{$\times$}}

\subsection{Experimental Setup}

\runin{Dataset}
We train and evaluate Drive2Gauss on the nuScenes benchmark~\cite{caesar2020nuscenesmultimodaldatasetautonomous}, which contains 1,000 urban driving scenes with six surround-view camera views. Following~\cite{wang2022readyvisioncentricdrivingstreaming}, we upsample the scene annotations, including 3D bounding boxes and road map information, from 2~Hz to 12~Hz to align them with the corresponding video generation setting. We follow the official split, using 700 scenes for training and 150 scenes for comprehensive validation.

\runin{Metrics}
We use PSNR, SSIM~\cite{wang2004imagequality}, LPIPS~\cite{zhang2018deepfeatures}, FID~\cite{heusel2017ttur}, and FVD~\cite{unterthiner2018fvd} to comprehensively evaluate both the 
generated and rendered results. Higher PSNR and SSIM values are better, whereas lower
LPIPS, FID, and FVD are preferred.
Unless otherwise noted, the best and second best results in the
 comparison tables are clearly marked with \bestlabel and \secondlabel, respectively.

\runin{Implementation Details}
Both components are trained on eight NVIDIA H20 GPUs in total. For multimodal generation, we initialize the generator from the pretrained DiST-T checkpoint~\cite{guo2025dist4ddisentangledspatiotemporaldiffusion} and train progressively at \(224\times400\) and \(424\times800\) resolutions. The training targets comprise RGB, metric depth, and dynamic flow, with the latter two constructed as described above in Sec.~\ref{sec:multimodal_generation}. For reconstruction, each decoder training sample contains four frames from three camera views at \(424\times800\). We apply \(k=2\) refinement iterations and optimize the decoder for four days
using AdamW with a learning rate of \(1\times10^{-3}\) and the corresponding image
reconstruction objective in Sec.~\ref{sec:query_decoder}, using loss weights
\(\lambda_p=0.2\), \(\lambda_o=0.001\), and \(\lambda_s=0.02\).

\subsection{Future Video Generation}

We evaluate future video generation along the original camera trajectories of
nuScenes. Each sample consists of 6 camera views and 17 consecutive video frames at
\(424\times800\) image resolution and a rate of 12~Hz. Given the six RGB images from the first
frame and the corresponding future control signals, Drive2Gauss is able to 
generate future RGB, metric depth, and dynamic flow jointly in a single pass.

\runin{Quantitative Comparison}
We evaluate RGB generation quality using the FID and FVD metrics, as reported in
Table~\ref{tab:future_video_generation}. Drive2Gauss obtains the lowest FID and
FVD among the listed methods and is the only one that jointly generates RGB,
metric depth, and dynamic flow in a unified model. Relative to DiST-T, which generates RGB and
depth without dynamic flow, FID decreases from 6.83 to 6.38 and FVD more than
halves from 22.67 to 12.43, indicating stronger temporal consistency. The
complete generator therefore retains high RGB quality while supplying the
geometric and motion cues that are required by 4D Gaussian reconstruction.

\begin{table}[!t]
\centering
\caption{Comparison of future video generation quality and output
modalities. Checkmarks indicate the modalities generated by each method.
Lower FID and FVD values are better.}
\label{tab:future_video_generation}
\scriptsize
\setlength{\tabcolsep}{5pt}
\begin{tabular}{lccccc}
\toprule
Method & RGB & Depth & Flow & FID $\downarrow$ & FVD $\downarrow$ \\
\midrule
Panacea~\cite{wen2023panaceapanoramiccontrollablevideo} & \cmark & \xmark & \xmark & 16.96 & 139.00 \\
MagicDrive~\cite{gao2024magicdrivestreetviewgeneration} & \cmark & \xmark & \xmark & 16.20 & 217.94 \\
Drive-WM~\cite{wang2024drivewm} & \cmark & \xmark & \xmark & 15.80 & 122.70 \\
WoVoGen~\cite{lu2024wovogen} & \cmark & \xmark & \xmark & 27.60 & 417.70 \\
Vista*~\cite{gao2024vistageneralizabledrivingworld} & \cmark & \xmark & \xmark & 13.97 & 112.65 \\
DriveDreamer-2~\cite{zhao2025drivedreamer2} & \cmark & \xmark & \xmark & 11.20 & 55.70 \\
MagicDriveDiT~\cite{gao2025magicdrivev2highresolutionlongvideo} & \cmark & \xmark & \xmark & 20.91 & 94.84 \\
UniScene~\cite{li2025unisceneunifiedoccupancycentricdriving} & \cmark & \xmark & \xmark & \secondcell{6.45} & 71.94 \\
DiST-T~\cite{guo2025dist4ddisentangledspatiotemporaldiffusion} & \cmark & \cmark & \xmark & 6.83 & \secondcell{22.67} \\
\textbf{Drive2Gauss} & \cmark & \cmark & \cmark & \bestcell{6.38} & \bestcell{12.43} \\
\bottomrule
\end{tabular}
\end{table}

\begin{figure}[t]
\centering
\includegraphics[width=\columnwidth]{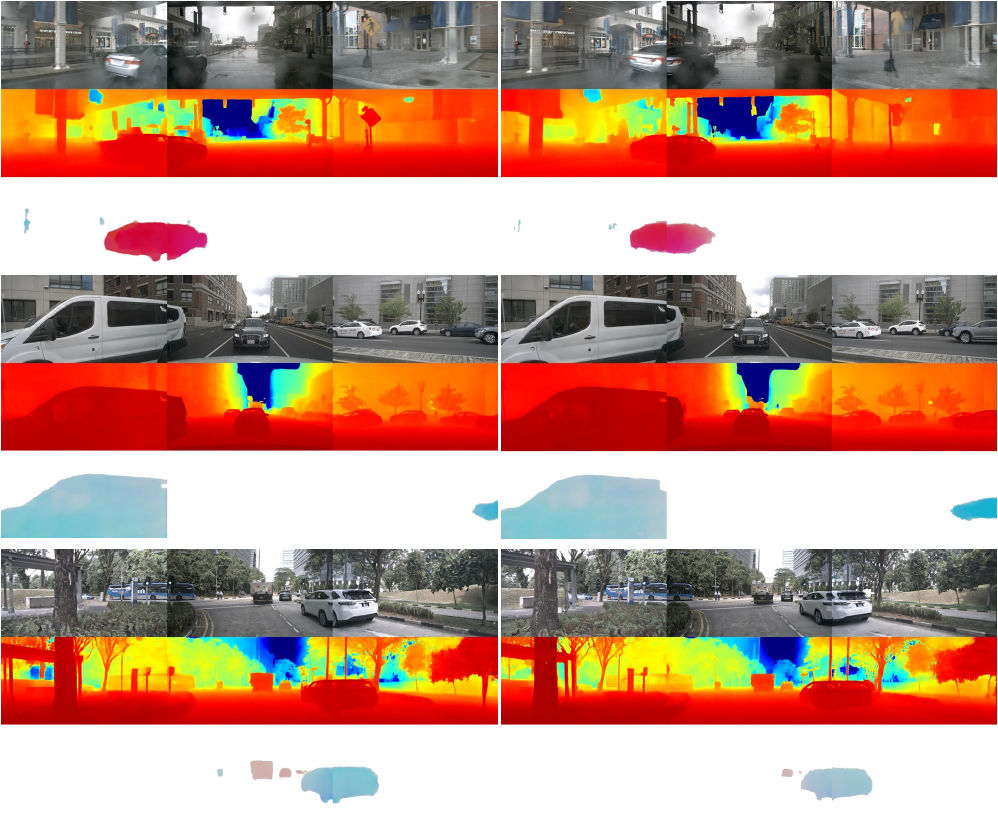}
\caption{
Qualitative results of joint multimodal future generation.
For each example, we visualize the generated RGB, metric depth, and dynamic flow
from the front three camera views of each sample. The three modalities together provide complementary
appearance, geometric, and motion information for the subsequent 4D Gaussian
construction and rendering.
}
\label{fig:multimodal_generation}
\end{figure}

\runin{Qualitative Comparison}
The three generated modalities from the selected camera views are visualized in
Fig.~\ref{fig:multimodal_generation}. The RGB outputs preserve the scene layout
and vehicle appearance, while depth recovers the relative geometry of the road,
vehicles, and background. The dynamic flow highlights the moving participants and
provides temporal correspondences for the Gaussian Query Decoder module in our framework.

\subsection{Feedforward 4D Gaussian Reconstruction}
\label{sec:reconstruction}

\runin{Quantitative Comparison}
We evaluate feedforward 4D Gaussian reconstruction on the sequences from the
preceding generation experiment. With zero-based frame indices, the four resulting 
reconstruction windows are \( [0,3] \), \( [4,7] \), \( [8,11] \), and
\( [12,15] \). They cover frames \(0\)--\(15\) of each 17-frame sequence; frame
\(16\) lies outside the range of these reconstruction windows. Each window supplies four
consecutive frames from the three front-facing input cameras.

For the RGB generation baselines, we use the same DGGT~\cite{dggt2025feedforward4dreconstruction}
model, initialized from its Waymo pretrained weights and finetuned on nuScenes,
to reconstruct 4D Gaussian scenes from their generated RGB videos. Drive2Gauss
instead directly reconstructs the scene from its generated RGB, metric depth,
and dynamic flow, so the comparison covers complete pipelines with different
inputs and decoders.

Table~\ref{tab:reconstruction_rendering} reports the highest reconstruction PSNR
and SSIM and the lowest LPIPS, FID, and FVD for Drive2Gauss among all the evaluated
pipelines. Its PSNR of 29.91~dB measures fidelity to its own generated RGB
observations. PSNR, SSIM, and LPIPS use each pipeline's generated observations
as the reference, whereas FID and FVD compare the rendered outputs with original
nuScenes videos. Together, these metrics describe input reconstruction fidelity
and rendered video quality; they do not separately attribute the improvements to the
generator and decoder components.

\begin{table}[!t]
\centering
\caption{Quantitative comparison of feedforward 4D Gaussian reconstruction.
PSNR, LPIPS, and SSIM are computed against the generated RGB observations,
while FID and FVD are computed against the original nuScenes videos.}
\label{tab:reconstruction_rendering}
\scriptsize
\setlength{\tabcolsep}{3pt}
\renewcommand{\arraystretch}{1.02}
\begin{tabular}{@{}lccccc@{}}
\toprule
Method & PSNR $\uparrow$ & LPIPS $\downarrow$ & SSIM $\uparrow$ &
FID $\downarrow$ & FVD $\downarrow$ \\
\midrule

Panacea + DGGT~\cite{wen2023panaceapanoramiccontrollablevideo,dggt2025feedforward4dreconstruction}
& 26.23 & 0.261 & 0.790 & 50.24 & 181.51 \\

MagicDrive-V2 + DGGT~\cite{gao2025magicdrivev2highresolutionlongvideo,dggt2025feedforward4dreconstruction}
& \secondcell{28.12} & 0.230 & \secondcell{0.830} & 68.09 & 370.59 \\

DiST-4D + DGGT~\cite{guo2025dist4ddisentangledspatiotemporaldiffusion,dggt2025feedforward4dreconstruction}
& 27.76 & \secondcell{0.200} & 0.825
& \secondcell{30.17} & \secondcell{90.70} \\

\textbf{Drive2Gauss}
& \bestcell{29.91} & \bestcell{0.101} & \bestcell{0.884}
& \bestcell{12.85} & \bestcell{32.01} \\

\bottomrule
\end{tabular}
\end{table}

\begin{figure*}[t]
\centering
\includegraphics[width=0.9\textwidth]{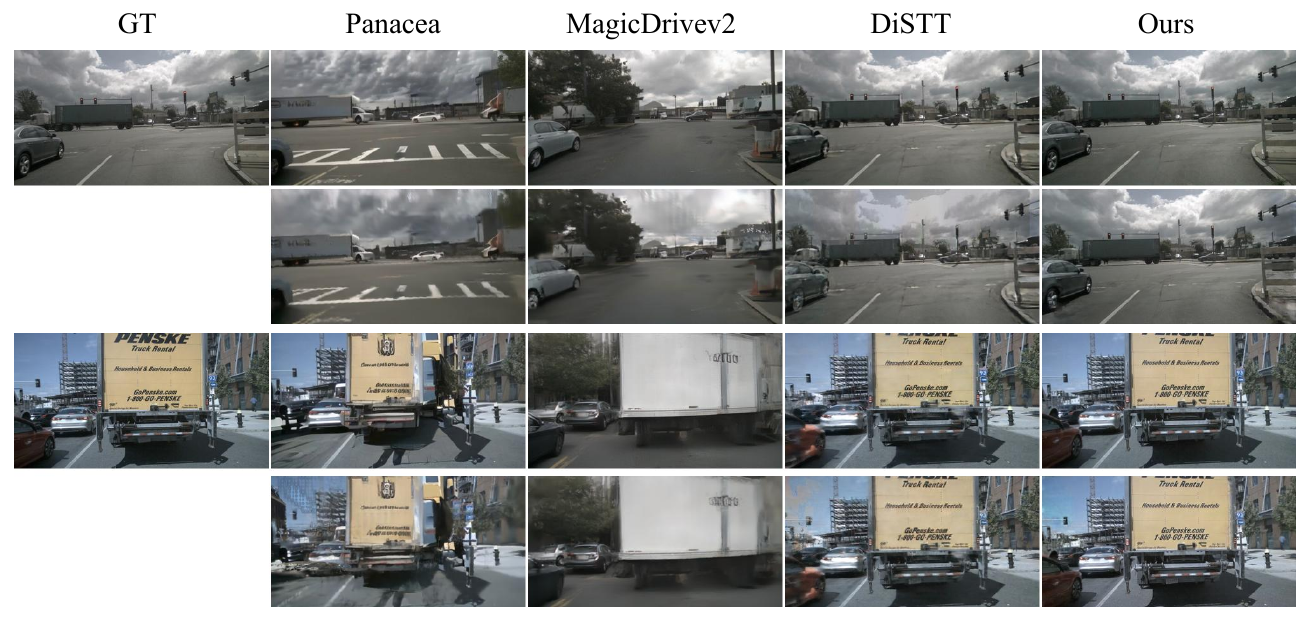}
\caption{
Qualitative comparison of feedforward 4D Gaussian reconstruction.
For each method, the upper row shows the generated RGB observation used as
input, while the lower row shows the corresponding rendering from the
reconstructed 4D Gaussian scene. The comparison highlights how well each
method preserves scene layout, object appearance, and local geometric
details between inference and rendering.
}
\label{fig:infer_render_compare}
\end{figure*}

\runin{Qualitative Comparison}
Each generated RGB observation is compared directly in
Fig.~\ref{fig:infer_render_compare} with the rendering obtained from its corresponding 
reconstructed 4D Gaussian scene. The paired rows expose differences in road boundaries, vehicle contours, and lane
markings. Drive2Gauss maintains closer correspondence, whereas RGB-only
pipelines show more blur and visible discrepancies around dynamic scene objects.

\subsection{Novel View Synthesis}

\runin{Quantitative Comparison}
Following FreeVS~\cite{wang2024freevsgenerativeviewsynthesis}, we evaluate
novel view synthesis under lateral camera shifts of
\(\tau \in \{\pm1\,\mathrm{m},\,\pm2\,\mathrm{m},\,\pm4\,\mathrm{m}\}\).
We use the same four reconstruction windows and three front-facing input views
as in Sec.~\ref{sec:reconstruction}. The front camera is shifted laterally
while its intrinsics, orientation, and timestamps remain unchanged. We evaluate
the rendered results using FID and FVD relative to the nuScenes video
distribution. RGB baselines use the same finetuned DGGT model, whereas
Drive2Gauss directly constructs the scene from the generated RGB, depth, and flow;
rendering is evaluated on the \texttt{CAM\_FRONT} view.

Drive2Gauss achieves the lowest FID at all the evaluated shifts and the lowest FVD
at $\pm1$~m and $\pm2$~m (Table~\ref{tab:lateral_shift_infer}). At $\pm1$~m, its
FVD is 281.27, compared with 562.22 for the DiST-T + DGGT baseline. At $\pm4$~m, however, its
FVD of 1025.44 exceeds the 956.98 obtained by DiST-T + DGGT, which indicates weaker
video distribution quality by this metric at the largest shift. These
distributional metrics primarily assess rendered visual quality rather than the pointwise
geometric accuracy at the shifted camera viewpoints.

\begin{table}[!t]
\centering
\caption{Direct novel-view synthesis under lateral camera shifts from generated
future observations. Lower FID and FVD are better.}
\label{tab:lateral_shift_infer}
\scriptsize
\setlength{\tabcolsep}{1.5pt}
\renewcommand{\arraystretch}{1.08}
\begin{tabular}{@{}l*{6}{c}@{}}
\toprule
Method & \multicolumn{2}{c}{$\pm1$ m} & \multicolumn{2}{c}{$\pm2$ m} &
\multicolumn{2}{c}{$\pm4$ m} \\
\cmidrule(lr){2-3} \cmidrule(lr){4-5} \cmidrule(lr){6-7}
& FID $\downarrow$ & FVD $\downarrow$
& FID $\downarrow$ & FVD $\downarrow$
& FID $\downarrow$ & FVD $\downarrow$ \\
\midrule
DreamDrive~\cite{mao2025dreamdrivegenerative4dscene}
& 72.20 & \secondcell{381.93}
& 143.15 & 780.71
& 233.15 & 1271.42 \\

Panacea + DGGT~\cite{wen2023panaceapanoramiccontrollablevideo,dggt2025feedforward4dreconstruction}
& 142.32 & 664.29
& 190.57 & 844.70
& 251.93 & 1085.29 \\

MagicDrive-V2 + DGGT~\cite{gao2025magicdrivev2highresolutionlongvideo,dggt2025feedforward4dreconstruction}
& 86.40 & 1032.22
& 109.34 & 1241.29
& 150.05 & 1481.17 \\

DiST-T + DGGT~\cite{guo2025dist4ddisentangledspatiotemporaldiffusion,dggt2025feedforward4dreconstruction}
& \secondcell{53.44} & 562.22
& \secondcell{73.55} & \secondcell{744.32}
& \secondcell{107.93} & \bestcell{956.98} \\

\textbf{Drive2Gauss}
& \bestcell{38.38} & \bestcell{281.27}
& \bestcell{59.60} & \bestcell{605.65}
& \bestcell{101.23} & \secondcell{1025.44} \\
\bottomrule
\end{tabular}
\end{table}

\begin{figure*}[t]
\centering
\includegraphics[width=0.9\textwidth]{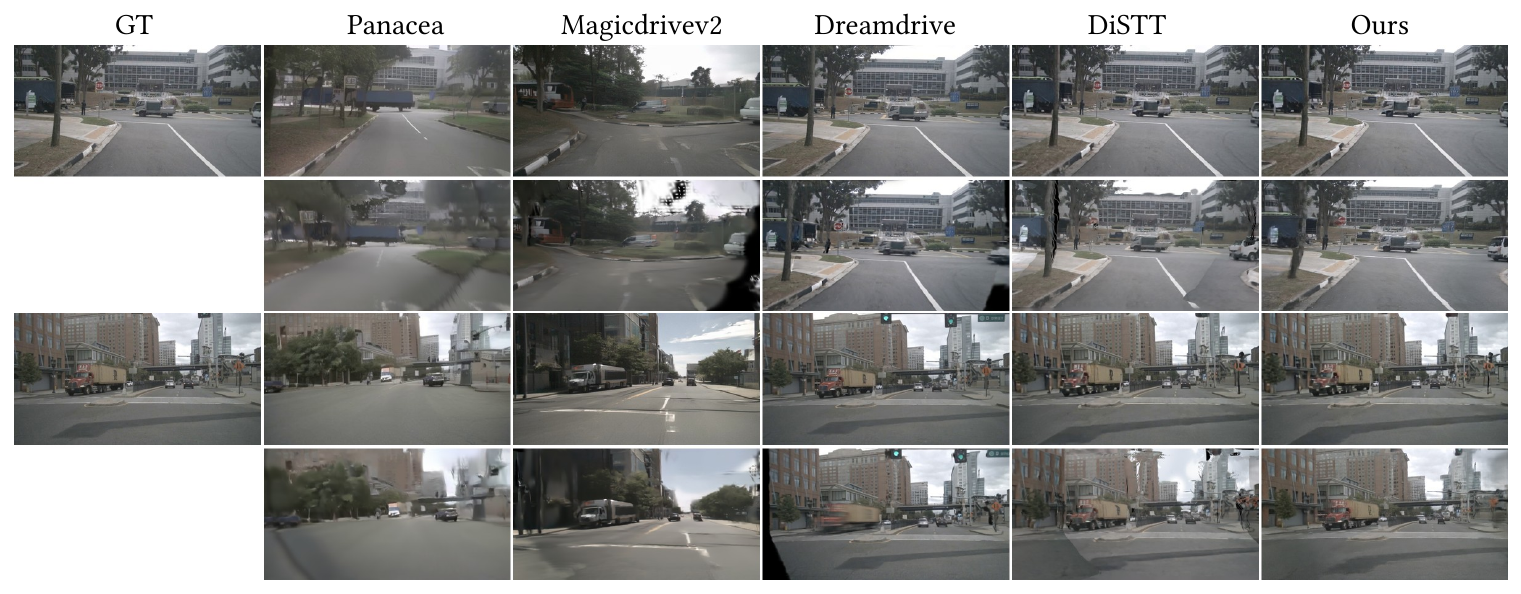}
\caption{
Qualitative comparison of novel-view rendering under lateral camera shifts.
For each scene, the upper row shows the generated or ground-truth observation,
while the lower row shows the corresponding rendering from the reconstructed
4D Gaussian scene.
}
\label{fig:novel_render_compare}
\end{figure*}

\runin{Qualitative Comparison}
Renderings after lateral translations of the front camera are compared directly in
Fig.~\ref{fig:novel_render_compare}. At small shifts, Drive2Gauss preserves the lane
boundaries, vehicle contours, and nearby background structures. With larger
shifts, all methods degrade, but Drive2Gauss retains more stable geometry; the
remaining artifacts expose the sensitivity to generated geometry and scene
coverage quality.

\subsection{Application to View Fixing}

\runin{Quantitative Comparison}
We further evaluate whether the reconstructed 4D Gaussian scenes can be reused
by a downstream view-fixing method. We apply SPVC~\cite{li2026spvc} to
Drive2Gauss and introduce an RGB-D projection baseline that directly lifts the
generated RGB and depth into Gaussian primitives before applying the same SPVC 
module. Table~\ref{tab:lateral_shift_fix} also lists alternative baseline pipelines.
These methods operate on real observations, whereas Drive2Gauss reconstructs
scenes from generated multimodal future observations.

Drive2Gauss + SPVC obtains the lowest FID and FVD in
Table~\ref{tab:lateral_shift_fix}. Relative to RGB-D + SPVC, it improves both
metrics at every shift, supporting the use of learned Gaussian
reconstruction as input to video fixing. These scores characterize the combined
reconstruction and SPVC pipeline; direct Gaussian rendering is evaluated
separately in Table~\ref{tab:lateral_shift_infer}.

\begin{table}[!t]
\centering
\caption{View-fixing performance under lateral camera shifts.
Lower FID and FVD are better.}
\label{tab:lateral_shift_fix}
\scriptsize
\setlength{\tabcolsep}{1.5pt}
\renewcommand{\arraystretch}{1.08}
\begin{tabular}{@{}l*{6}{c}@{}}
\toprule
Method &
\multicolumn{2}{c}{$\pm1$ m} &
\multicolumn{2}{c}{$\pm2$ m} &
\multicolumn{2}{c}{$\pm4$ m} \\
\cmidrule(lr){2-3}
\cmidrule(lr){4-5}
\cmidrule(lr){6-7}
&
FID $\downarrow$ & FVD $\downarrow$ &
FID $\downarrow$ & FVD $\downarrow$ &
FID $\downarrow$ & FVD $\downarrow$ \\
\midrule

DiST-S~\cite{guo2025dist4ddisentangledspatiotemporaldiffusion}
& 10.12 & 45.14
& 12.97 & 68.80
& 17.57 & 105.29 \\

GaussianDWM~\cite{deng2026gaussiandwm}
& 8.36 & 44.50
& 11.27 & 68.17
& 18.81 & 116.40 \\

WorldSplat~\cite{zhu2025worldsplatgaussiancentricfeedforward4d}
& \secondcell{8.25} & 40.17
& 11.26 & 47.41
& 13.38 & 64.07 \\

RGB-D + SPVC~\cite{li2026spvc}
& 10.37 & \secondcell{25.99}
& \secondcell{11.19} & \secondcell{30.68}
& \secondcell{12.82} & \secondcell{41.29} \\

\textbf{Drive2Gauss + SPVC}~\cite{li2026spvc}
& \bestcell{6.68} & \bestcell{21.25}
& \bestcell{7.85} & \bestcell{24.28}
& \bestcell{9.00} & \bestcell{35.73} \\

\bottomrule
\end{tabular}
\end{table}

\begin{figure}[t]
\centering
\includegraphics[width=\columnwidth]{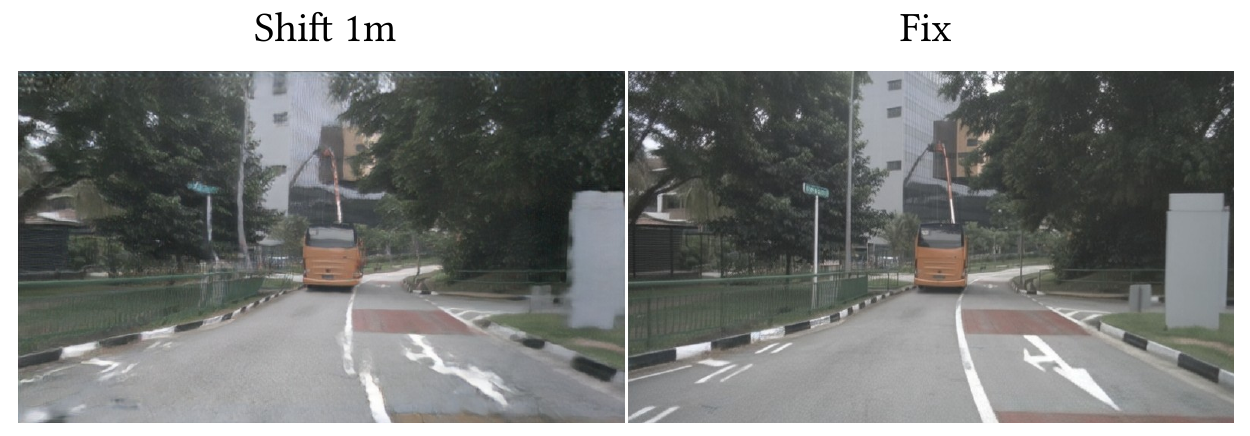}
\caption{Qualitative example of the downstream view fixing pipeline in our framework. Applying SPVC to the
Drive2Gauss reconstruction substantially reduces the artifacts introduced by the lateral camera shift.}
\label{fig:view_fixing_example}
\end{figure}

\runin{Qualitative Comparison}
The effect of applying SPVC after a lateral camera shift is illustrated in
Fig.~\ref{fig:view_fixing_example}. The shifted rendering distorts the road markings
and nearby vehicle, while view fixing restores local road structure and
foreground alignment. This supports using the reconstructed scene as an
intermediate representation for downstream correction.

\subsection{Ablation Studies}

\runin{Component Ablation}
We ablate the three design choices of the Gaussian Query Decoder. All variants
share the same generated observations, query budget, four-frame three-view
context, and training schedule; only the component under study is modified. 
For \emph{w/o Init Flow}, we set the flow-derived temporal displacement
to zero during query initialization, retaining the geometric anchors
provided by source-frame metric depth.
For \emph{w/o Cross-Attention}, we replace cross-attention with an
MLP-based weighted fusion module that predicts observation-wise weights for
feature aggregation. For \emph{w/o Iterative Refinement}, we remove the iterative
position update and feature resampling after initial aggregation stage.

\begin{table}[!t]
\centering
\caption{Ablation study of the Gaussian Query Decoder. Metrics follow the
reconstruction protocol of Table~\ref{tab:reconstruction_rendering}.}
\label{tab:decoder_ablation}
\scriptsize
\setlength{\tabcolsep}{7pt}
\renewcommand{\arraystretch}{1.05}
\begin{tabular}{@{}lcc@{}}
\toprule
Variant & PSNR $\uparrow$ & SSIM $\uparrow$ \\
\midrule
w/o Init Flow
& 27.89 & 0.814 \\

w/o Iterative Refinement
& 28.03 & 0.827 \\

w/o Cross-Attention
& 28.76 & 0.849 \\

\textbf{Drive2Gauss (Full)}
& \textbf{29.91} & \textbf{0.884} \\
\bottomrule
\end{tabular}
\end{table}
Table~\ref{tab:decoder_ablation} shows that all three choices contribute: removing
any one of them lowers both PSNR and SSIM, most severely for the generated flow and
least for cross-attention, so flow-guided query association and iterative
refinement of query positions matter more than specific aggregation
operator.

\runin{Query Number Sensitivity}
We further study how the number of Gaussian queries affects reconstruction quality
and cost, varying only the number of initialized queries per frame-view while every
other setting remains completely as above. Table~\ref{tab:query_sensitivity} reports the
resulting quality together with the training peak memory and the inference time per sample.
Quality improves with the query budget, but the PSNR gain shrinks as the budget
grows while SSIM continues to improve, so additional queries mainly refine
structural fidelity at the upper end of the studied query range considered here. Training memory grows
roughly linearly with the query budget in our setting and the largest configuration exceeds the
memory available on a single H20 GPU, whereas inference time grows only
sublinearly. Memory rather than inference cost is therefore the main binding
constraint, and we use 960K queries in all main experiments by default.

\begin{table}[!t]
\centering
\caption{Sensitivity to the number of Gaussian queries. Training peak memory
is reported in GB, while inference time is reported in seconds.}
\label{tab:query_sensitivity}
\scriptsize
\setlength{\tabcolsep}{4pt}
\renewcommand{\arraystretch}{1.08}
\begin{tabular}{@{}lcccc@{}}
\toprule
Queries & PSNR $\uparrow$ & SSIM $\uparrow$ &
\shortstack{Train Peak\\Memory (GB)} &
\shortstack{Inference\\Time (s)} \\
\midrule
480K & 28.87 & 0.811 & 45 & 0.96 \\
720K & 29.62 & 0.836 & 67 & 1.11 \\
\textbf{960K} & \textbf{29.91} & \textbf{0.884} & \textbf{92} & \textbf{1.29} \\
1200K & / & / & \textsc{OOM} & / \\
\bottomrule
\end{tabular}
\end{table}

\section{CONCLUSION}

In this paper, we present Drive2Gauss, a framework for feedforward
generative driving scene reconstruction. Its Multimodal Video Generator jointly predicts RGB, metric
depth, and dynamic flow. These complementary predictions guide the
Gaussian Query Decoder to construct reusable 4D Gaussian scenes
through depth-based initialization and flow-guided aggregation,
without per-scene optimization. Experiments on nuScenes demonstrate
state-of-the-art performance in video generation, 4D reconstruction, and
view fixing, with the lowest FID across all evaluated lateral camera shifts.

\bibliographystyle{IEEEtran}
\bibliography{references}

\end{document}